%% file: ijcai25.tex
\documentclass{article}
\usepackage{ijcai25}

\usepackage{times}
\usepackage{soul}
\usepackage{url}
\usepackage[hidelinks]{hyperref}
\usepackage[utf8]{inputenc}
\usepackage[small]{caption}
\usepackage{graphicx}
\usepackage{amsmath}
\usepackage{amsthm}
\usepackage{booktabs}
\usepackage{algorithm}
\usepackage{algorithmic}
\usepackage[switch]{lineno}
\usepackage{xcolor}
\usepackage{fontawesome}
\usepackage{amssymb}
\usepackage{mydef}
\usepackage{tikz}
\usepackage[edges]{forest}

\title{Human-Centric Foundation Models: Perception, Generation \\ and Agentic Modeling}

\author{
Shixiang Tang$^1$
\and
Yizhou Wang$^1$\and
Lu Chen$^{2}$\and
Yuan Wang$^3$\and
Sida Peng$^2$\and
Dan Xu$^4$\and
Wanli Ouyang$^1$\\
\affiliations
$^1$The Chinese University of Hong Kong~~~
$^2$State Key Lab of CAD\&CG, Zhejiang University \\
$^3$Tsinghua University~~~
$^4$Hong Kong University of Science and Technology
}

\begin{document}

\maketitle

\begin{abstract}

Human understanding and generation are critical for modeling digital humans and humanoid embodiments. Recently, Human-centric Foundation Models (HcFMs)—inspired by the success of generalist models such as large language and vision models—have emerged to unify diverse human-centric tasks into a single framework, surpassing traditional task-specific approaches. In this survey, we present a comprehensive overview of HcFMs by proposing a taxonomy that categorizes current approaches into four groups: (1) Human-centric Perception Foundation Models that capture fine-grained features for multi-modal 2D and 3D understanding; (2) Human-centric AIGC Foundation Models that generate high-fidelity, diverse human-related content; (3) Unified Perception and Generation Models that integrate these capabilities to enhance both human understanding and synthesis; and (4) Human-centric Agentic Foundation Models that extend beyond perception and generation to learn human-like intelligence and interactive behaviors for humanoid embodied tasks. We review state-of-the-art techniques, discuss emerging challenges and future research directions. This survey aims to serve as a roadmap for researchers and practitioners working towards more robust, versatile, and intelligent digital human and embodiments modeling. \textcolor{blue}{\href{https://github.com/HumanCentricModels/Awesome-Human-Centric-Foundation-Models/}{Website is available.}}
\end{abstract}

\input{fig/treemap}

\section{Introduction}
Recent years have witnessed remarkable strides towards understanding human appearance, emotions, identities, actions, intentions and generating photorealistic humans in 2D and 3D. The success of these methods can be attributed to the robust estimation of identification~\cite{he2024instruct,li2024all}, 2D keypoints~\cite{wang2023hulk,yuan2024hap}, fine-grained body-part segmentation~\cite{tang2023humanbench,chen2023beyond}, depth~\cite{khirodkar2024sapiens}, textual descriptions~\cite{chen2024language}, and human meshes~\cite{cai2024smpler}, as well as the powerful human-centric deep learning frameworks,  \emph{e.g.,} vision transformers~\cite{jin2024you,wang2023hulk,huang2024refhcm} and diffusion models~\cite{ju2023humansd,li2024cosmicman,lin2025omnihuman1}. Despite progress in every individual task, robust and accurate understanding and generating photorealistic and even intelligent digital humans requires to deeply understand human as a holistic and complex system at the intersection of diverse human-centric tasks related to appearance, identity, motion and intentions. Moreover, most existing human-centric pipelines are task-specific for better performances, leading to huge costs in representation/network design, pretraining, parameter-tuning, and annotations. Therefore, the recent human-centric learning community appeals for a unified framework~\cite{ci2023unihcp,wang2023hulk,chen2024language,huang2024refhcm} to unlock systematic understanding and a wide range of human-centric applications for everybody.

Inspired by rapid advancements of general foundation models, \emph{e.g.,} large language models (LLMs), large vision models (LVMs) and text-to-image generative models, and their presents of a paradigm shift from end-to-end learning of task-specific models to generalist models, a recent trend is to develop \textbf{H}uman-\textbf{c}entric \textbf{F}oundation \textbf{M}odels \textbf{(HcFM)} that satisfy three criteria, namely generalization, broad applicability, and high fidelity. Generalization ensures robustness to unseen conditions, enabling the model to perform consistently across varied environments. Broad applicability indicates the versatility of the model, making it suitable for a wide range of tasks with minimal modifications, or even without modifications. High fidelity denotes the ability of the model to produce precise, high-resolution outputs essential for faithful human generation tasks such as 2D to 3D lifting. Recent notable works towards such human-centric foundation models include SOLIDER~\cite{chen2023beyond}, PATH~\cite{tang2023humanbench}, UniHCP~\cite{ci2023unihcp}, Sapines~\cite{khirodkar2024sapiens}, MotionGPT~\cite{jiang2023motiongpt,zhang2024motiongpt}, ChatHuman~\cite{lin2024chathuman}, \emph{etc}.

In light of the rapid developments and emerging challenges of Human-centric foundation models, we present a comprehensive survey of this field to help the community keep track of its progress. Specifically, we introduce a taxonomy that categorizes existing works into four groups according to their supported downstream tasks: \emph{Human-centric perception foundation models}, \emph{Human-centric AIGC foundation models}, \emph{Human-centric unified perception and generation foundation models}, and \emph{Human-centric agentic foundation models}. Human-centric perception foundation models modify the existing unsupervised and multitask supervised pretraining framework to capture the fine-grained human-centric features essential to multi-modal 2D and 3D perception tasks, such as skeleton-based action recognition, human parsing, \emph{etc}. Human-centric AIGC foundation models build on the success of generative foundation models by leveraging rich human-specific data, focusing on creating human-related content with high realism and diversity. To move a step forward, human-centric perception and generation tasks can be unitedly modeled in the single foundation model with the inspiration of multi-modal large language models. Such unified modeling are proven not only to benefit human understanding but also fine-grained and photorealistic generation. Lastly, as mentioned by Michael J Black and Xavier Puig~\cite{yao-feng-no-date}, the human foundation agent model goes beyond perception and generation, it is a model that can receive signals from human sensors, and interact with humans based on the inputs and, for which, the embodiment, actions, and motor behaviors are human-like. Different from other foundation models, the human-centric agent foundation models aim at learning human intelligence, which hopefully can benefit various embodied AI tasks.

To our best knowledge, this is the first survey about human-centric foundation models with a novel taxonomy (Fig.~\ref{taxonomy}). Its scope also extends beyond the mere categorization of existing techniques. It also explores the potential future trajectories, contemplating how ongoing advancements might unfold, including data, ethics and technological aspects.


\begin{figure*}[t]
    \centering
    \includegraphics[width=0.99\linewidth]{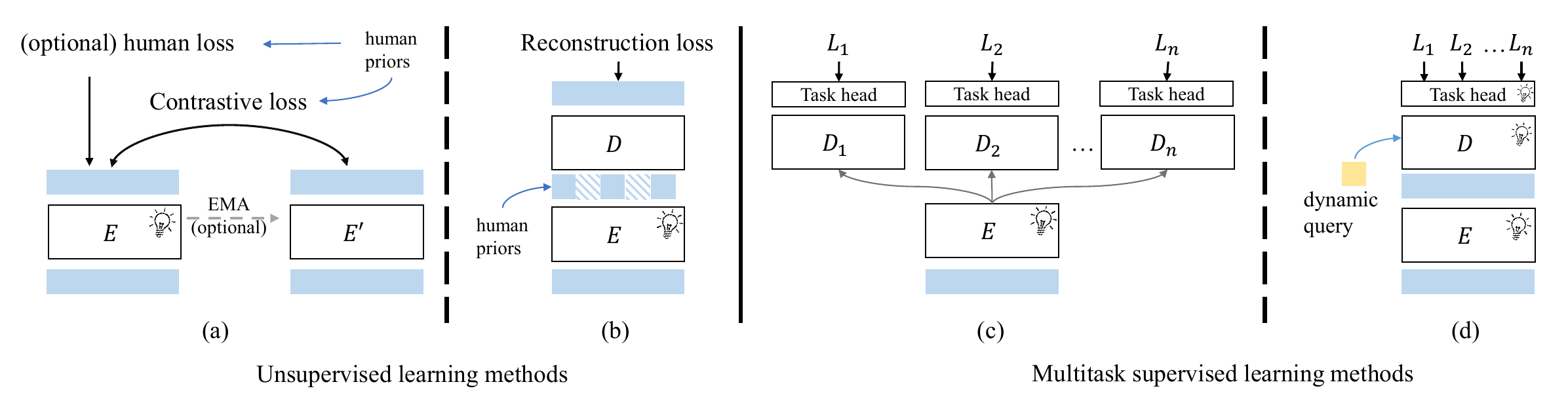}
    \vspace{-0.75em}
    \caption{Different Frameworks of human-centric perception foundation models. Parameters in modules with \faLightbulbO ~are used in downstream tasks. Unsupervised foundation models: (a) Contrastive learning methods; (b) Mask image modeling methods. Supervised foundation models: (c) Multitask supervised pretraining methods; (d) Unified modeling methods.}
    \label{fig:perception}
    \vspace{-0.5em}
\end{figure*}

\section{Taxonomy}
The objective of the taxonomy is to group Human-centric foundation models with the similar supported tasks into the same category. Specifically, we classify existing human-centric foundation models into four categories, \emph{i.e.,} Human-centric perception foundation models, Human-centric AIGC foundation models, Human-centric unified perception and generation foundation models, and Human-centric agentic models, which can be further summarized in difference learning frameworks. We briefly introduce the four categories as follows:

\noindent \textbf{(1) Human-centric perception foundation models} learn from large-scale and multi-modal human-centric data and support major perceptive tasks, including person re-identification, human parsing, pose estimation, human mesh recovery and skeleton-based action recognition. The fundamental idea to leverage structures of human bodies or diverse annotations to learn the fine-grained and semantic human-centric representations. Based on the learning frameworks, human-centric perception foundation models can be further categorized into unsupervised learning and multitask supervised learning.

\noindent \textbf{(2) Human-centric AIGC foundation models} are designed to create content—such as images, videos, or avatars—that centers on human. These models are trained on extensive human-centric data to produce realistic and diverse portrayals of individuals, whose objective is to produce content that accurately reflects fine-grained human appearances, behaviors, and interactions. Based on the training paradigm, these models can be broadly classified into models trained with unsupervised learning and multi-modal supervised learning.

\noindent \textbf{(3) Human-centric unified perception and generation foundation models} can support perception and generation tasks closely correlated. The fundamental idea is to view human-centric cues other than texts as foreign languages and append them to large language models (LLMs) as multi-modal large language models (MLLM) for understanding and generation. 

\noindent\textbf{(4) Human-centric agentic foundation models} learn human intelligence and support human-centric embodied AI tasks, such as human-robot collaboration tasks and social interactions. Based the learning frameworks, human-centric agentic foundation models can be categorized into vision-language-based method and vision-language-action-based methods.

The four categories of foundation models are interconnected rather than mutually exclusive, contributing collectively to the rapid advancement of the human-centric foundation models. In the follows, we delve into each of them, exploring key challenges, representative solutions, and emerging trends.


\section{Human-centric Perception Foundation Models} \label{sec:perception}
The human-centric perception foundation model demonstrates that compact ahuman-centric representations can be learned  from multiple human-centric task and efficiently adapted to a broad range of 2D and 3D multi-modal perception tasks. Based on whether to leverage human-centric annotations, we cover two paradigms of human-centric perception foundation models: unsupervised learning and supervised learning.

\subsection{Unsupervised learning methods}
Human-centric unsupervised foundation models were proposed to mitigate dependency on the sensitive annotations, which mainly follow the pretraining-finetuning paradigm. During pretraining, without labels, they used the inherent priors of human body structure to learn versatile and representative human-centric features in the encoder. When adapted to downstream tasks, the pretrained encoder and the task head will be parameter-efficiently and fully finetuned using labeled data.

\noindent{\textbf{Contrastive learning methods}} leverage human priors to align features derived from encoders using 
tailored contrastive losses, as shown in Fig.~\ref{fig:perception}(a). 
Considering the multi-modal nature of human data (\emph{e.g.}, RGB, depth, 2D keypoints), instead of commonly used momentum encoders, multiple encoders were used to process inputs with different modalities. Human priors were included to guide multi-modal contrastive learning for general human-centric representations.
HCMoCo~\cite{hong2022versatile} employed multiple encoders to exploit multi-modal human body consistency through a hierarchical contrastive learning framework. Based on it, PBoP~\cite{meng2024efficient} introduced an additional encoder for generated latent part pair images. The extracted features could then serve as anchors to guide the multi-modal contrasting learning process. However, the acquisition of multi-modal data is not always straightforward. 
When only images were available, integrating an additional loss with human prior knowledge was another effective solution.
For example, SOLIDER~\cite{chen2023beyond} proposed an additional semantic classification loss to import semantic information into the learned features. LiftedCL~\cite{chen2023liftedcl} introduced an adversarial loss to supervise the lifted 3D skeletons,
explicitly inserting 3D human structure information for human-centric pretraining.

\noindent\textbf{Mask image modeling methods} implicitly learn human knowledge by reconstructing masked inputs (see Fig.~\ref{fig:perception}(b)) based on the prior knowledge of body structures. HAP~\cite{yuan2024hap} utilized the 2D keypoints to guide the mask sampling process during mask image modeling, encouraging the model to concentrate on body structure information. To introduce 3D human prior, ~\cite{armando2024cross} proposed to reconstruct masked pedestrian images from cross-view and cross-pose pairs. 
Thanks to human priors,  human-centric unsupervised foundation models show superior performance than the ImageNet pretraining methods on human-centric perception tasks, especially on low-data regimes.

\subsection{Multitask supervised learning}
When labled data is sufficient, multitask supervised learning can exploit the internal relationship among data, emerging as a straightforward and effective paradigm for constructing human-centric perception foundation models. 
Some recent works learned the shared information among different human-centric datasets 
to benefit specific human-centric tasks,
\emph{e.g.}, human shape estimation~\cite{cai2024smpler}, pedestrian detection~\cite{zhang2024pedestrian}, re-identification~\cite{he2024instruct,li2024all}. 
By constructing a unified framework applicable to related subtasks and training them concurrently, these approaches have outperformed specialist models.
However, a notable limitation of these methods is their inability to learn and execute other perception tasks, which restricts the potential scope of the application.
In response to the challenge, recent advances in multiple human-centric tasks co-training showed a new direction. These developments demonstrate that human-centric perception foundation models can 
exploit inter-task homogeneity to enhance overall performance. 

\noindent\textbf{Multitask supervised pretraining methods} centered on utilizing multiple distinct supervisions to force the encoder to learn general human-centric representation (see Fig.~\ref{fig:perception}(c)). To handle task conflicts brought by supervised learning from diverse labels, 
PATH~\cite{tang2023humanbench} leveraged task-specific projectors along with the hierarchical weight-sharing strategy, enforcing the encoder to acquire general representations for downstream human-centric tasks. 

\noindent\textbf{Unified modeling methods} have emerged as the mainstream in human-centric perception foundation models to further mitigate the resource-intensive full finetuning process for downstream tasks. 
As depicted in Fig.~\ref{fig:perception}(d), these methods followed the unified encoder-decoder framework with dynamic queries.
UniHCP~\cite{ci2023unihcp}, as a first attempt, adopted a task-guided interpreter to unify task heads and task-specific queries, handling five human-centric tasks. HQNet~\cite{jin2024you} focused on instance-level features for individual persons and proposed Human Queries to learn unified all-in-one query representations, 
providing a single-stage method to tackle multiple distinctive human-centric tasks.
While these works mainly concentrate on 2D human-centric tasks, Hulk~\cite{wang2023hulk} extended the scope to simultaneously address 2D vision, 3D vision, vision-language and skeleton-based human-centric tasks. To tackle these tasks, Hulk categorized the input and output formats into four modalities and developed modality-specific (de-)tokenizers with modality indicator queries, unifying all tasks into modality translation tasks. 
Considering human-computer interaction, RefHCM~\cite{huang2024refhcm} converted multi-modal data into semantic tokens, unifying various perception tasks as referring tasks. 




\section{Human-centric AIGC Foundation Models} \label{sec:aigc}
\input{sections/04_AIGC}

\section{Human-centric Unified Perception and Generation Foundation Models}  \label{sec:unified}
In recent years, human-centric foundation models emerge as a transformative approach for unifying perception and generation tasks, providing comprehensive frameworks to understand and synthesize human behaviors, motions, emotions, and intentions. By integrating multi-modal human-centric cues—such as visual, auditory, textual, emotional, and motion data—into LLMs and MLLMs, these models facilitate richer, context-aware representations of human interactions. Such models can be categorized into two primary paradigms—\textit{fixed vocabulary} and \textit{extended vocabulary}-based on how multi-modal human-centric signals are integrated into LLMs and MLLMs.




\subsection{Fixed Vocabulary}

Vocabulary-Fixed models enhance LLMs by introducing modality-specific projection layers to map human-centric signals into the feature space of LLMs or directly employing off-the-shelf tools and prompt engineering technique. Among these approaches, CoMo~\cite{huang2024controllable} unified the text conditioned human motion generation, fine-grained motion generation and motion editing. Specifically, it auto-regressively generates sequences of interpretable pose codes upon the high-level text description and fine-grained, body-part-specific descriptions generated by LLMs. ChatPose~\cite{feng2024chatpose} introduced LLMs to advance pose-related tasks, striving to develop a versatile pose generator. By integrating image interpretation, world knowledge, and body language comprehension into foundational LLMs, ChatPose enhanced its ability to understand and reason about 3D human poses from images and textual descriptions. Taking this a step further, ChatHuman~\cite{lin2024chathuman} introduced an multi-modal LLM integrated with 22 domain-specific human-centric tools, improving its ability to reason about human-related tasks. Assisted by academic publications and retrieval-augmented generation model, ChatHuman generated in-context-learning examples for handling newly introduced tools. In other human-centric tasks, foundational models have also made remarkable strides. For example, ChatGarment~\cite{bian2024chatgarment} leveraged large vision-language models (VLMs) to automate the estimation, synthesis, and editing of 3D garments from either images or textual descriptions. Similarly, FaceGPT~\cite{wang2024facegpt} integrated 3D morphable face model (3DMM)~\cite{blanz2023morphable} parameters into token space of VLMs, enabling the self-supervised generation of 3D facial models from both textual and visual inputs.

\begin{figure}
    \centering
    \includegraphics[width=0.99\linewidth]{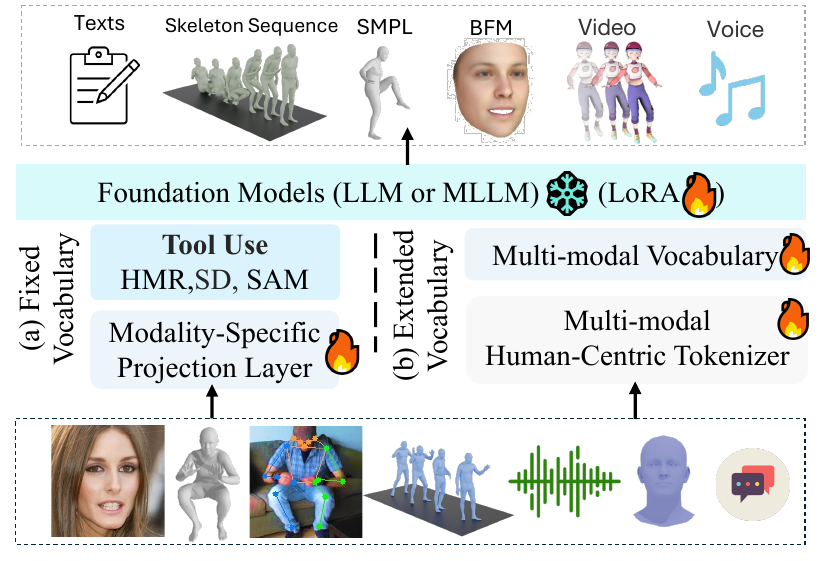}
    \caption{Human-centric foundation models for unified perception and generation tasks. (1) Vocabulary-Fixed models with fixed vocabulary introduces modality-specific projection layers or directly employing off-the-shelf human-centric tools. (2) Vocabulary-extended models align enriched human-centric representations, \textit{e.g.}, pose parameters, SMPL representations, motion sequences, emotions, and audio signals with the original vocabulary of foundational LLMs.}
    \label{fig:unfied}
\end{figure}

\subsection{Extended Vocabulary} 
Vocabulary-extended models are designed to expand capabilities of LLMs~\cite{touvron2023llama,achiam2023gpt} and MLLMs~\cite{liu2024visual,li2023blip} by explicitly extending their vocabulary and embedding spaces to accommodate human-centric and multi-modal signals. By aligning enriched human-centric representations—such as \textit{pose parameters}, \textit{SMPL representations}, or \textit{motion sequences}—with the original vocabulary of LLMs, Vocabulary-Extension Models empower foundational LLMs and MLLMs to effectively tackle novel human-centric tasks, including generating, editing, and comprehending human poses of textual descriptions, image, and 3D modalities.
Recent studies~\cite{wu2024motionllm,wang2024motiongpt,jiang2023motiongpt,luo2024m,zhou2024avatargpt} incorporated motion modality into textual space of the foundational LLMs, enabling a holistic representation of the complex relationship between motion and natural language. MotionGPT~\cite{jiang2023motiongpt} introduced a unified motion-language model to handle multiple motion-related tasks. It first discretized continuous motions into discrete semantic tokens, which could be interpreted as ``\textit{body language}'' and then be extended to the LLM's vocabulary. Through pre-training alignment and prompt tuning stages, MotionGPT showcased strong generalization across various human motion understanding and generation tasks. Building on the MotionGPT~\cite{jiang2023motiongpt}, AvatarGPT~\cite{zhou2024avatargpt} proposed an \textit{all-in-one} structure for motion understanding, planning, generation, as well as motion in-between
synthesis. M3-GPT~\cite{luo2024m} further embeded motion, music, and language into a single vocabulary and includes music-to-dance and dance-to-music tasks. Such a unified approach advanced human-centric tasks by synthesizing multi-modal human behaviors and bridging the gap between auditory, visual, and linguistic modalities. While these methods mostly were restricted to single-human motion, MotionLLM~\cite{wu2024motionllm} introduced a simple yet versatile framework capable of handling single-human, multi-human motion generation, and motion captioning by fine-tuning pre-trained LLMs.
MotionGPT-2~\cite{wang2024motiongpt} developed a general-purpose Large Motion-Language Model (LMLM), which extended beyond current solutions by tackling the challenging 3D holistic motion generation on the MotionX~\cite{lin2023motion} benchmark. However, these methods struggled with pose-related editing and comprehension. UniPose~\cite{li2024unipose} leveraged generation abilities of LLMs to unify all pose-relevant tasks to comprehend, generate, and edit human poses across various modalities, including images, text, and 3D SMPL representations. As a solution, LOM~\cite{chen2024language} unified verbal and non-verbal language using MLLMs for human motion understanding and generation, accommodating text, speech, motion, or any combination thereof as input flexibly. It trained a compositional body motion VQ-VAE to tokenize motions into part-ware discrete tokens, unifying modality-specific vocabularies (audio and text).

\section{Human-centric Agentic Foundation Models}  \label{sec:agent}

\begin{figure}[t]
    \centering
    \includegraphics[width=1.01\linewidth]{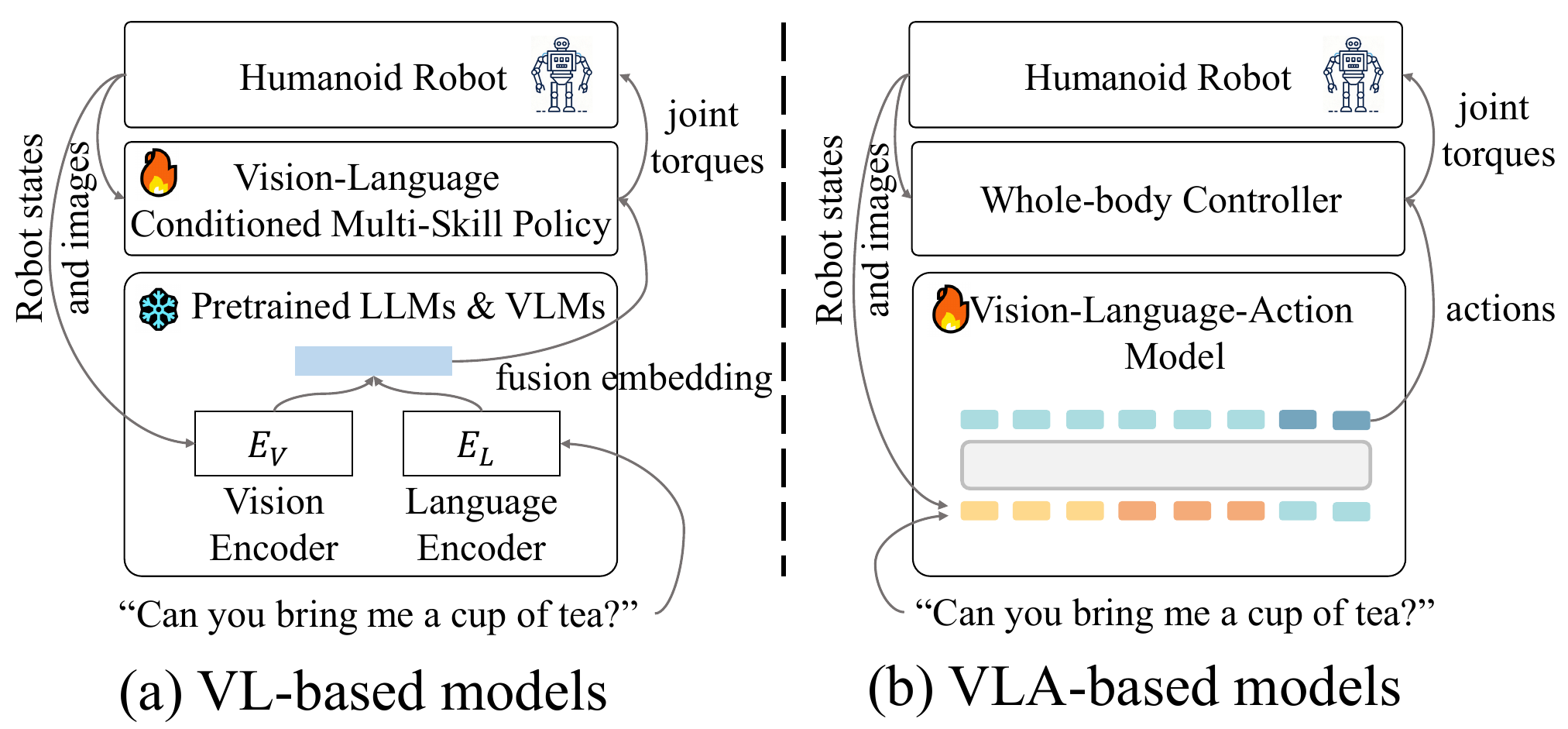}
    \caption{Frameworks of human-centric agentic foundation models to endow humanoid robots. (a) Vision-Language-based models: Use pretrained vision-language to learn multi-skill policy; (b) Vision-Language-Action-based models: Finetune a vision-language-action model to generate actions.}
    \label{fig:rfm_framework}
\end{figure}

Beyond understanding and generating human themselves, Human-centric agentic foundation models often process mulitmodal inputs (\emph{e.g.,} egocentric images, tactiles, sounds, and natural language as task description) focus on the human reaction and interactions to the environments while being constrained by its physical properties. Leveraging the Internet-scale multi-modal dataset, Human-centric agentic foundation models hold the promise of generalization across diverse tasks and provide a natural interface for human-robot interaction, both essential for real-world robot applications.

\subsection{Vision-language-based models}
Vision-language-based Models are prominent approaches in constructing foundation models for humanoid systems by integrating pre-trained vision-language models (VLMs) with sensorimotor control policies. As illustrated in Fig.~\ref{fig:rfm_framework}(a), these methods leverage the strengths of visual and linguistic modalities to bridge high-level semantic understanding with low-level physical actions.
In these frameworks, visual inputs are processed using pre-trained image encoders while natural language commands are tokenized through pre-trained text encoders. The extracted semantic features are then aligned with humanoid-specific control policies, enabling the translation of language and visual cues into precise motor actions. This cross-modal alignment not only facilitates language-to-action mapping but also enhances the capability of humanoid robots to perform complex, multi-step tasks. Recent efforts have focused on adapting these vision-language-based models to the unique challenges of humanoid robotics. Notable examples include HumanVLA~\cite{xu2024humanvla} and SuperPADL~\cite{juravsky2024superpadl}, which aligned humanoid control policies with the latent spaces of pre-trained VLMs. These approaches have demonstrated the potential to endow humanoid robots with sophisticated skills that are driven by image and language inputs, paving the way for more advanced and natural human-robot interactions in dynamic environments.

\subsection{Vision-language-action-based models}
Vision-language-action (VLA) models have emerged as a promising approach in constructing foundation models for humanoid robots by unifying visual, linguistic, and action modalities within a single framework. As depicted in Fig.~\ref{fig:rfm_framework}(b), these methods treat robot data—both observations and actions—as tokens in the vocabulary of a pre-trained language model, thereby enabling direct fine-tuning or co-training with established vision-language models. In contrast to traditional methods that depend on separately trainable low-level control policies, VLA models directly generate actions as token sequences, effectively merging high-level semantic reasoning with motor command generation. This unified tokenization framework allows the model to generate a wide range of skills while preserving robust language and vision understanding, thus enhancing its generalization across diverse tasks. However, representing actions as stringified tokens may become inefficient for high degrees-of-freedom systems such as humanoid robots, where the complexity of motor commands is significantly higher. Despite the potential benefits, applying the VLA paradigm to humanoid robotics remains largely unexplored. To address this gap, recent initiatives like NVIDIA's Project GR00T~\cite{dong2024bringing} have been announced. The GR00T foundation model aspired to leverage a diverse array of data sources—from internet and simulation data to real-robot interactions—to facilitate scalable training and achieve robust, cross-modal performance for complex humanoid tasks. 


\section{Challenges and Future Directions}

\noindent \textbf{Data.} Different from general images and videos, collecting high-quality human-centric data is much more sensitive, difficult and expensive, compared to general images and videos. This situation inevitably leads to a trade-off in data quantity and data quality. Furthermore, the wide variability in human appearances, behaviors and contexts makes it hard to get a comprehensive dataset, 
limiting the generalization ability of data-driven human-centric foundation models. 

\noindent \textbf{Representations.} Human appearance and behavior demand a holistic understanding that integrates body, face, and hands, yet no existing foundation model captures these aspects simultaneously. Recognizing that these elements are closely interrelated parts of a complex system, there remains a significant gap in unified modeling frameworks. Future research should focus on developing innovative, multi-modal architectures and scalable datasets that bridge the global and granular representations of human dynamics, ultimately enhancing applications in digital human synthesis, interactive robotics, and personalized human-computer interaction.


\noindent \textbf{Interactivity.} Despite significant progress in understanding and generating isolated human attributes such as appearance, emotion, and identity, current human-centric foundation models face challenges in capturing the complex interplay of interactions and contextual dynamics inherent in real-world scenarios. Most approaches are limited to static or isolated representations, hindering their ability to model the nuanced interdependencies between individuals and their environments. Future research should focus on developing unified frameworks that seamlessly integrate high-level semantic reasoning with fine-grained behavioral synthesis, enabling models to adapt to dynamic, multi-agent contexts.

\noindent \textbf{Ethics.} Ethics are crucial in applying human-centric foundation models, especially regarding sensitive domains and privacy. 
In sensitive domains, the model's output should never replace the expertise of human professionals.
Moreover, Human-centric foundation models should also be developed with privacy-enhancing technologies, ensuring that the model learns powerful representations from the data while making it difficult to identify individual-level privacy information. 
Additionally, anonymization methods should be applied to all training data to avoid privacy violations.

\clearpage
\bibliographystyle{named}
\bibliography{ijcai25}

\end{document}

%% file: fig/treemap.tex
\tikzstyle{leaf}=[draw=hiddendraw,
    rounded corners,minimum height=1em,
    fill=hidden-orange!40,text opacity=1, align=center,
    fill opacity=.5,  text=black,align=left,font=\scriptsize,
    inner xsep=3pt,
    inner ysep=2.5pt,
    ]
\begin{figure*}[ht]
\centering
\begin{forest}
  for tree={
  forked edges,
  grow=east,
  reversed=true,
  anchor=base west,
  parent anchor=east,
  child anchor=west,
  base=middle,
  font=\scriptsize,
  rectangle,
  draw=hiddendraw,
  rounded corners,align=left,
  minimum width=2em,
    s sep=5pt,
    inner xsep=3pt,
    inner ysep=2.5pt,
  },
  where level=1{text width=4.5em}{},
  where level=2{text width=6em,font=\scriptsize}{},
  where level=3{font=\scriptsize}{},
  where level=4{font=\scriptsize}{},
  where level=5{font=\scriptsize}{},
  [Human-centric Foundation Models,rotate=90,anchor=north,edge=hiddendraw
    [Perception (\S\ref{sec:perception}),edge=hiddendraw,align=center,text width=5.7em
        [Self-supervised Learning, text width=9.2em, edge=hiddendraw
            [Contrastive Learning,leaf,text width=6.7em, edge=hiddendraw
                        [SOLIDER~\cite{chen2023beyond}{,} HCMoCo~\cite{hong2022versatile}{,}\\
                        PBoP~\cite{meng2024efficient}{,}
                        LiftedCL~\cite{chen2023liftedcl},
                        leaf,text width=21em, edge=hiddendraw]
                        ]
            [Mask Image Modeling,leaf,text width=6.7em, edge=hiddendraw
                        [HAP~\cite{yuan2024hap}{,} Sapiens~\cite{khirodkar2024sapiens},
                        leaf,text width=21em, edge=hiddendraw]
                        ]
        ]
        [Multi-task Supervised Learning, text width=9.2em, edge=hiddendraw
            [Multi-task Supervised Pretraining,leaf,text width=9.7em, edge=hiddendraw
                        [PATH~\cite{tang2023humanbench},leaf,text width=18em, edge=hiddendraw]
            ]
            [Unified Modeling,leaf,text width=9.7em, edge=hiddendraw
                        [UniHCP~\cite{ci2023unihcp}{,} HQNet~\cite{jin2024you}{,}\\
                        Hulk~\cite{wang2023hulk}{,}
                        RefHCM~\cite{huang2024refhcm},leaf,text width=18em, edge=hiddendraw]
                                    ]
        ]
    ]
    [AI-Generated \\Content (\S\ref{sec:aigc}),edge=hiddendraw,align=center,text width=5.7em
     [Unsupervised Learning, text width=6.8em, edge=hiddendraw
            [GANs with Style Modulation,leaf,text width=8.6em, edge=hiddendraw
            [StyleGAN-Human~\cite{fu2022stylegan}{,} UnitedHuman~\cite{fu2023unitedhuman},leaf,text width=21.5em, edge=hiddendraw]
            ]
            [GANs with Neural Renderer,leaf,text width=8.6em, edge=hiddendraw
            [AniPortraitGAN~\cite{wu2023aniportraitgan}{,} AG3D~\cite{dong2023ag3d},leaf,text width=21.5em, edge=hiddendraw]
            ]
      ]
    [Multi-modal Supervised \\Learning, text width=6.8em, edge=hiddendraw
            [Conditional Latent Diffusion Models,leaf,text width=11.8em, edge=hiddendraw
                [HumanSD~\cite{ju2023humansd}{,} HyperHuman~\cite{liu2023hyperhuman}{,} \\CosmicMan~\cite{li2024cosmicman} ,leaf,text width=18.3em, edge=hiddendraw]
                        ]
            [Spatial-Temporal Diffusion Transformers,leaf,text width=11.8em, edge=hiddendraw
                [Human4DiT~\cite{shao2024360}{,} OmniHuman~\cite{lin2025omnihuman1} ,leaf,text width=18.3em, edge=hiddendraw]
                        ]
        ]
    ]
    [Unified Perception \\
    \& Generation (\S\ref{sec:unified}), edge=hiddendraw,align=center,text width=5.7em
      [Fixed Vocabulary,text width=6.2em, edge=hiddendraw
            [CoMo~\cite{huang2024controllable}{,} ChatPose~\cite{feng2024chatpose}{,} ChatHuman~\cite{lin2024chathuman}{,} \\ChatGarment~\cite{bian2024chatgarment}{,} FaceGPT~\cite{wang2024facegpt},leaf,text width=32.5em, edge=hiddendraw]
      ]
      [Extended Vocabulary,text width=6.2em, edge=hiddendraw
            [MotionGPT~\cite{jiang2023motiongpt}{,} M3-GPT~\cite{luo2024m}{,} UniPose~\cite{li2024unipose}{,} MotionLLM\\\cite{wu2024motionllm}{,} MotionGPT-2~\cite{wang2024motiongpt}{,} AvatarGPT~\cite{zhou2024avatargpt}{,} LOM~\cite{chen2024language}, leaf,text width=32.5em, edge=hiddendraw]
      ]
    ]
    [Agentic Model (\S\ref{sec:agent}), edge=hiddendraw,align=center,text width=5.7em
      [Vision-Language-based Models,text width=11.3em, edge=hiddendraw
            [HumanVLA~\cite{xu2024humanvla}{,}
            SuperPADL~\cite{juravsky2024superpadl},leaf,text width=19.5em, edge=hiddendraw]
      ]
      [Vision-Language-Action-based Models,text width=11.3em, edge=hiddendraw
            [Roject GR00T~\cite{dong2024bringing},leaf,text width=19.5em, edge=hiddendraw]
      ]
    ]
    ]
\end{forest}
\caption{A taxonomy of human-centric foundation models with representative examples.}
\label{taxonomy}
\vspace{-0.5em}
\end{figure*}

%% file: sections/04_AIGC.tex
Human-centric AIGC foundation models are generative models specifically trained or fine-tuned on \textit{large-scale human datasets} to create content focused on human elements, such as images, videos, and 3D avatars. By prioritizing the generation fidelity of human attributes, these models serve as \textit{adaptable bases for downstream tasks} like 
image editing, virtual try-on, 3D generation, and character animation. 
In this section, we categorize these models based on unsupervised learning and multi-modal supervised learning approaches.

\subsection{Unsupervised Learning}

Unsupervised learning plays a crucial role in early human-centric generative models, with GAN-based methods leading the progress in realistic human image and avatar generation. Specifically, there are two main frameworks: \textit{2D GANs with Style Modulation} and \textit{3D-Aware GANs with Neural Renderer}.



\begin{figure*}[t]
    \centering
    \includegraphics[width=0.99\linewidth]{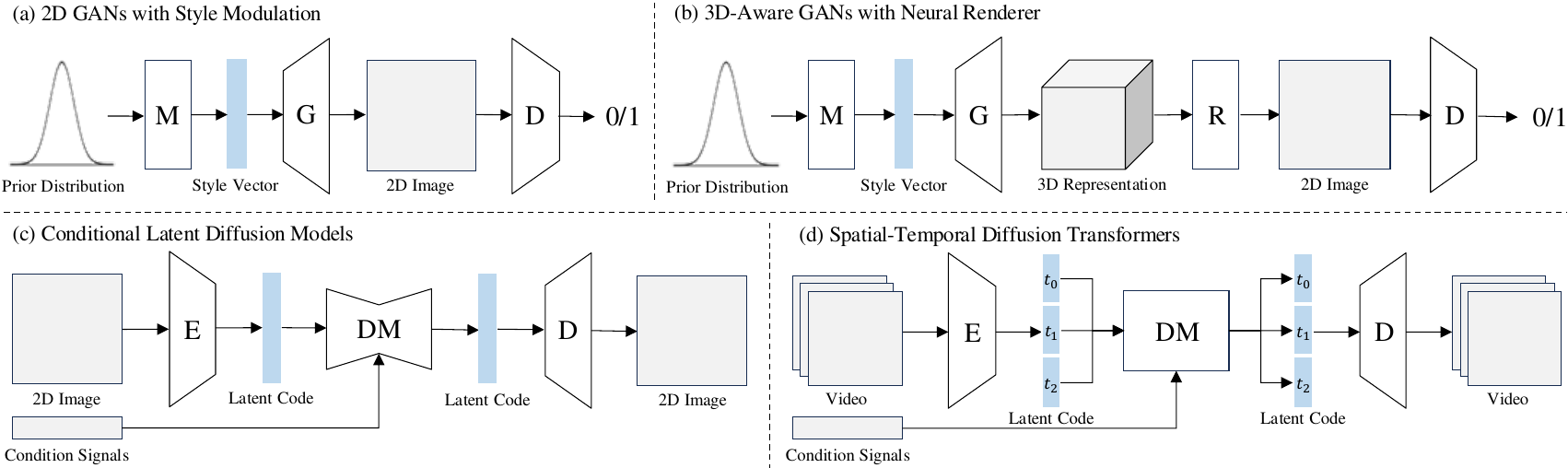}
    \vspace{-0.5em}
    \caption{Different frameworks of human-centric AIGC foundation models. Unsupervised learning methods: (a) A sampled noise is transformed by a mapping network (M) into a style code that modulates the generator (G) at multiple scales, while a discriminator (D) is trained simultaneously through adversarial learning to refine realism. (b) 3D representations are incorporated along with a neural renderer (R) to enhance spatial consistency. Supervised learning methods: (c) The input image is encoded (E) into a latent code, which the diffusion model (DM) iteratively denoises to reconstruct high-quality outputs with fine-grained control. (d) The input video is compressed and decomposed into space-time latents, where transformers capture spatial-temporal dependencies and enable efficient scaling for video generation.}
    \label{fig:aigc}
     \vspace{-0.5em}
\end{figure*}

\noindent{\textbf{GANs with Style Modulation}} leverage an intermediate, disentangled latent space to exert fine-grained control over the image synthesis process by first mapping a random noise to a style vector and then using learned affine transformations to inject style vectors at various layers of the generator, as shown in Fig.\ref{fig:aigc} (a).
This hierarchical modulation allows fine‐grained control of visual attributes at different scales.  For instance, StyleGAN-Human~\cite{fu2022stylegan} trained a series of unconditional models with this framework and showed that scaling data size, balancing data distribution, and aligning human body could significantly enhance generation performance, thereby establishing robust foundational models for conditional applications. UnitedHuman~\cite{fu2023unitedhuman} employed a multi-source spatial transformer to align multi-source data, including face, hand, partial-body, and whole-body images, into a unified space for more comprehensive human modeling, and designed a continuous generator to synthesize coherent high-relsolution images with enhanced details. These models served as versatile, pre‐trained foundations for a variety of downstream applications.
Their well-structured latent space enabled (1) intuitive drag manipulation of image attributes~\cite{pan2023drag} by optimizing latent codes based on point movement supervision,
(2) 3D generation~\cite{xiong2023get3dhuman} with reconstruction priors to produce disentangled geometry and texture code, and (3) virtual try-on~\cite{yoshikawa2023stylehumanclip} by aligning the latent code with garment features.

\noindent{\textbf{GANs with Neural Renderer}} integrate 3D representations and a neural renderer into the former framework to enforce 3D geometric consistency during image synthesis, (Fig.\ref{fig:aigc}(b)). By guiding the generator with rendering results (e.g., low-resolution image, depth, or surface normals), these models produce outputs that better reflect realistic spatial structures, making them especially to create high-fidelity 3D avatars. For example,
SofGAN~\cite{chen2022sofgan} decoupled the 3D representation space into geometry and texture subspaces, providing a robust foundation for independent control over camera pose, facial structure, and attribute texture in various human portraits. 
AniPortraitGAN~\cite{wu2023aniportraitgan} integrated human priors into its framework by learning pose and facial deformations using the SMPL model and 3D Morphable Model (3DMM).
Trained on large-scale facial image collections, its generator and renderer could serve as the basis for tasks that require high-resolution controllable images with detailed 3D geometry.
Similarly, AG3D~\cite{dong2023ag3d} extended this framework to train on large-scale full-body images by incorporating an additional deformer for learning pose-dependent effects and a normal estimator for geometry supervision, resulting in high-quality 3D human avatars. These models' 3D-aware generation ability unlocked many downstream tasks, including (1) free-viewpoint video synthesis by rendering from arbitrary camera trajectories, (2) pose retargeting
by integrating learnable pose-dependent deformations, and (3) single-view 3D reconstruction with GAN inversion.



\subsection{Multi-modal Supervised Learning}

Recent advances in diffusion models have spurred the development of multi-modal supervised learning approaches that leverage large-scale paired datasets, such as text-image or video-pose alignments, to achieve precise control over the generation process.
With \textit{Conditional Latent Diffusion Models} and \textit{Spatial-Temporal Diffusion Transformer Architecture}, recent works have significantly improved the quality, consistency, and diversity of synthetic human images and videos.


\noindent{\textbf{Conditional Latent Diffusion Models}} extend standard diffusion models by operating in a compact latent space and incorporating external conditions, such as text, pose, or segmentation maps, to guide the generation process (Fig.\ref{fig:aigc} (c)). By first encoding inputs into a lower-dimensional latent representation, diffusion is applied to gradually denoise the output while preserving the provided conditions. This approach enhances both efficiency and controllability, making it a general framework for tasks like human image synthesis, pose-guided generation, and multi-modal content creation. Specifically,
HumanSD~\cite{ju2023humansd} proposed a skeleton-guided diffusion model with a heatmap-guided denoising loss. Trained on 2M+ text-image-pose triplets, this model demonstrated its foundational ability in generating high-quality human images across various scenarios.
HyperHuman~\cite{liu2023hyperhuman} presented a unified framework for generating hyper-realistic human images by capturing correlations between appearance and structure in multi-modal data. Specifically, it introduced a latent structural diffusion model that jointly denoises depth, surface normal, and RGB images conditioning on caption and pose skeleton, where modality-specific branches can be complementary to each other.
Recently, CosmicMan~\cite{li2024cosmicman} was proposed by building on three key pillars: a scalable, high-quality data production paradigm, robust model design via a decomposed attention-refocusing framework, and pragmatic integration into downstream tasks. This structure empowered CosmicMan to generate images in various scenarios, from full-body portraits to close-up shots, establishing it as a versatile cornerstone for human-centric content generation.



\noindent{\textbf{Spatial-Temporal Diffusion Transformers}} extend diffusion models by incorporating transformer-based architectures to capture both spatial structures and temporal dependencies in videos. By leveraging self-attention mechanisms across spatial dimensions and time steps, this framework (Fig.\ref{fig:aigc} (d)) effectively captures long-range temporal relationships while maintaining spatial consistency in human-centric video generation. Consequently, it is ideally suited for tasks such as character animation, video manipulation, and 4D generation. A representative work is Human4DiT~\cite{shao2024360}, which introduced a hierarchical 4D Diffusion Transformer (DiT) for generating high-quality, 360-degree spatial-temporally coherent human videos from a single image. Trained on a multi-sourced dataset spanning images, videos, multi-view captures, and 4D footages, the model factorized self-attention across views, time, and space while incorporating accurate condition injection, successfully handling complex motions and viewpoint changes.
Building on this framework, OmniHuman~\cite{lin2025omnihuman1} proposed a scalable, multi-modality-conditioned human video generation model from a single image and motion signals (e.g., audio, video, or both).
By leveraging mixed data and incorporating motion-related conditions during training, it mitigated data scarcity and enabled realistic video synthesis across diverse scenarios, including talking, singing, varying body compositions, and human-object interactions. Notably, pre-trained human-centric diffusion models are increasingly serving as the versatile foundation for a wide range of applications.
For example, they can be 
(1) fine-tuned for text-driven image editing~\cite{brooks2023instructpix2pix} using instruction-image pairs, 
(2) integrated into character animation~\cite{hu2024animate} by incorporating an additional branch to merge detailed textures into character's movements, 
(3) adapted for virtual try-on~\cite{karras2024fashion} by splitting classifier-free guidance into person, garment, and pose conditioning, and 
(4) extended to 3D/4D human generation~\cite{kolotouros2023dreamhuman} by iteratively applying score distillation sampling in multiple views.